# Variance-Adjusted Cosine Distance as Similarity Metric

## Satyajeet Sahoo[1] and Jhareswar Maiti[2]

## Abstract

Cosine similarity is a popular distance measure that measures the similarity between two vectors in the inner product space. It is widely used in many data classification algorithms like K-Nearest Neighbors, Clustering etc. This study demonstrates limitations of application of cosine similarity. Particularly, this study demonstrates that traditional cosine similarity metric is valid only in the Euclidean space, whereas the original data resides in a random variable space. When there is variance and correlation in the data, then cosine distance is not a completely accurate measure of similarity. While new similarity and distance metrics have been developed to make up for the limitations of cosine similarity, these metrics are used as substitutes to cosine distance, and do not make modifications to cosine distance to overcome its limitations. Subsequently, we propose a modified cosine similarity metric, where cosine distance is adjusted by variance-covariance of the data. Application of variance-adjusted cosine distance gives better similarity performance compared to traditional cosine distance. KNN modelling on the Wisconsin Breast Cancer Dataset is performed using both traditional and modified cosine similarity measures and compared. The modified formula shows 100% test accuracy on the data.

## 1 Introduction

Cosine Similarity is a popular similarity measure used in many applications, particularly in text mining and information retrieval to find similar documents or text because of its advantages over Euclidean Distance in handling high dimensional vectors of tokens, ease of computation and better performance in handling sparse data. and scale invariance. However, cosine similarity suffers from certain limitations e.g. underperforming when the vectors are unnormalized [1], providing less strict comparison as it does not take the magnitude of vectors into account and focuses only on the orientation [2] [3], being biased by high-value features without taking the number of shared features in two vectors into consideration [4] etc. In this study, we present another limitation i.e. cosine distance does not accurately measure similarity, if the data presents significant covariance/correlation. As data is represented in the form of a vector of random variables, in this study we refer the space in which the data is represented (with the random variables as coordinate axes) as the *random variable space*. In this study we show that since the random variable space is different to the Euclidean space in that in the Euclidean space data is present as a spheroidal distribution whereas in random variable space data distribution has non-spheroidal properties owing to the covariance structure, hence the operations that are valid in a cartesian coordinate space, like calculating the Euclidean distance and cosine distance, are not valid in a random variable space.

As the prevailing practice in various ML models involves applying cosine distance as a measure of similarity directly in the random variable space, it carries risk of misclassification. Hence to account for the non-spheroidal nature of the distribution of input data in order to use cosine distance as a true measure if similarity, we present a modified cosine distance formula. In this formula we have utilized the first principles behind Mahalanobis distance [5] [6] to obtain a transformation matrix by performing Cholesky decomposition of the covariance matrix of the data, which is then applied on the individual data points to transform the data from the random variable space to the Euclidean space by removing the variance and correlation effects. We show that the transformed space is a Euclidean space where data distribution being spheroidal means operations of cartesian coordinate space like Euclidean distance are applicable, hence by generalization cosine distance is also applicable in this space. We demonstrate this applicability by applying the original cosine distance and the modified cosine distance on the Wisconsin Breast Cancer Dataset in the KNN model [7] and compare the results. It is observed that the modified formula that adjusts for the variance and covariance/correlation effects provides 100% test accuracy. The approach of applying cosine similarity in transformed spaces has been considered in multiple studies e.g. Nguyen and Bai [8] computed cosine similarity for facial verification in a transformed subspace by learning a linear transformation A: $R^m \rightarrow R^d$. Goldberger et al [9] in their paper on Neighborhood Components Analysis have referred to the random variable space as input space and observed that KNN performs well when there is a linear transformation of the input space using inverse square root of positive semi-definite matrices. While they used the space transformation to develop a cost function based on stochastic neighbor assignments in the transformed space, the novelty of this study is that we have used the covariance matrix of the underlying distribution as the symmetric positive definite matrix and obtained the inverse square root using Cholesky decomposition and used it for transformation to

- *Satyajeet Sahoo is with Department of Industrial and Systems Engineering, IIT Kharagpur. E-mail: satyajeet.sahoo@ kgpian.iitkgp.ac.in.*
- *Prof J.Maiti is with Department of Industrial and Systems Engineering, IIT Kharagpur and is currently Chairman, CoE-Safety Engineering and Analytics, IIT Kharagpur E-mail: jmaiti@iem.iitkgp.ac.in*

develop a variance adjusted cosine distance that works in the transformed space.

The remainder of this paper is organized as follows: Section 2 presents the fallacies of using cosine distance in data which is presented as multivariate distributions in a random variable space. Section 3 presents the Cholesky factor which is then used to obtain variance adjusted cosine similarity metric. In Section 4 KNN on Wisconsin dataset is performed using both original and modified cosine formula. While studies have been conducted on application of variations of KNN on Wisconsin dataset like using multiple distance functions [10], or iterative weighted KNN to get missing feature values [11], the accuracy in these studies has varied from 85.71 to 98.85. In this study by performing data transformation, it is possible to get 100% test accuracy. In section 5 the results of both the methodologies are compared and the performance improvement of the modified formula is demonstrated. Finally, section 6 concludes with a summary of the contributions, future scope and future opportunities for model application.

## 2 LIMITATIONS OF COSINE DISTANCE

Cosine distance measures the similarity between two data points based on the following reasoning: two data points will show similarity if their respective vectors exhibit low angular distance. This is measured using the cosine of the angle between two vectors $X_i$ and $X_j$ and is given by the formula

$$\text{Cos}(\Theta) = \frac{X_i . X_j}{|X_i||X_j|}$$

where $X_i . X_j$ is the dot product and $|X_i|$ is the norm.

However, since the cosine distance does not take into account the nature of data distribution or the innate variance-covariance, this is valid when the data is distributed in a spheroidal fashion in the Euclidean space. Consider a non-spheroidal data distribution that exhibits unequal variance and covariance:

Fig.1: Contour of a bivariate Gaussian distribution

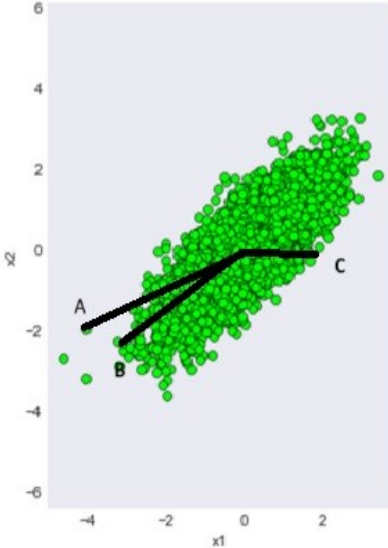

This is a bivariate Gaussian distribution on synthetic data with mean vector = 0. Three points A, B and C are selected and the angles between their vectors from the origin are compared. While angle between A and B is smaller compared to the angle between B and C, in reality, B exhibits greater similarity to C than to A. This is because B and C belong to the same distribution, whereas A is an outlier. Since the distribution is ellipsoid and not a spheroid, the covariance structure in the data impacts the distance between the points. Hence if we use vanilla cosine distance to find the similarity, then it will not give accurate results.

Hence the data needs to be decorrelated and whitened before cosine distance can be calculated. This is the basis of Mahalanobis distance, and the principles behind Mahalanobis distance have been used to obtain the transformation matrix which will adjust the cosine distance for data variance/covariance.

## 3 CHOLESKY DECOMPOSITION

Mahalanobis distance is represented as:

$$(X - \mu)^T \Sigma^{-1} (X - \mu) \tag{1}$$

Where $\Sigma_{p \times p}$ is the population covariance matrix with p variables.

Now using Cholesky whitening, since $\Sigma$ is a positive definite matrix, we get

$$\Sigma = \Lambda \Lambda^T$$

Where $\Lambda$ is a lower triangular matrix.

Then $\Sigma^{-1} = (\Lambda\Lambda^T)^{-1} = (\Lambda^T)^{-1}\Lambda^{-1}$

So, the equation then becomes

$$(X - \mu)^T \Sigma^{-1} (X - \mu) = (X - \mu)^T (\Lambda^T)^{-1}\Lambda^{-1}(X - \mu) \tag{2}$$

Now, $(\Lambda^T)^{-1} = (\Lambda^{-1})^T$

Also, we know $(AB)^T = B^T A^T$

The equation then becomes

$$(X - \mu)^T (\Lambda^T)^{-1}\Lambda^{-1}(X - \mu) = (X - \mu)^T (\Lambda^{-1})^T\Lambda^{-1}(X - \mu)$$

Or,

$$(X - \mu)^T (\Lambda^{-1})^T\Lambda^{-1}(X - \mu) = [\Lambda^{-1}(X - \mu)]^T[\Lambda^{-1}(X - \mu)] \tag{3}$$

Hence this can be represented as $Z^T Z$, where

$$Z = [\Lambda^{-1}(X - \mu)] = \Lambda^{-1}X - \Lambda^{-1}\mu \tag{4}$$

But $Z^T Z$ is the formula of the Euclidean distance. Hence Mahalanobis distance is essentially Euclidean distance between the data vectors in a new vector space to which the original data vector was transformed using transformation matrix $\Lambda^{-1}$, which is the inverse of Cholesky decomposed lower triangular matrix of the original covariance matrix. It means after vector transformation, distance between two data vectors in the new space can be measured using Euclidean distance formula. Hence if the original space where the data is present is the random variable space, the new vector space where Euclidean distance is valid is the Euclidean space and satisfies the spherical coordinate system. If the random variable vector in the original sample space is $X^{Random}$, then in the Euclidean space the transformed data vector will be given by

$$X^{Euclidean} = \Lambda^{-1} X^{Random}$$

Suppose we have a dataset where each data point belongs to one of the two class labels: y=1 or y=-1. If we consider that the class labels represent the two unique experiments which led to two distinct populations of datapoints being generated, then we can also say that the two class labels represent the two sample spaces corresponding to the experiments. Hence data points with class label y=1 will have population covariance matrix $\Sigma_{y=1}$. The corresponding lower triangular matrix for the covariance matrix will be $\Lambda_{y=1}$. Transforming the data points from this sample space to the Euclidean space leads to

$$X_{y=1}^{Euclidean} = \Lambda_{y=k}^{-1} X_{y=1}^{Random} \qquad (5)$$

Similarly, data points with class label y=-1 will have covariance matrix $\Sigma_{y=-1}$. The corresponding lower triangular matrix for the covariance matrix will be $\Lambda_{y=-1}$. Transforming the data points from this sample space to the Euclidean space leads to

$$X_{y=-1}^{Euclidean} = \Lambda_{y=k}^{-1} X_{y=-1}^{Random} \qquad (6)$$

In this study, we hypothesize **that since Euclidean distance (which is an operation of cartesian coordinate system) is not valid in the random variable space but is valid in the transformed Euclidean space, a generalization can be made that principles of the cartesian coordinate system are not valid in the original random variable space, but are valid in the new transformed Euclidean space. Hence cosine distance is applicable as a measure of similarity if data is transformed by applying the transformation matrix $\Lambda_{y=1}^{-1}$ or $\Lambda_{y=-1}^{-1}$.**

The modified cosine distance is then given by

$$Cosine\ (X^i, X^j) = \frac{\left(\Lambda_{y=k}^{-1} X_{y=k}^i\right) \cdot \left(\Lambda_{y=k}^{-1} X_{y=k}^j\right)}{\left|\Lambda_{y=k}^{-1} X_{y=k}^i\right| \left|\Lambda_{y=k}^{-1} X_{y=k}^j\right|} \qquad k \in \{0,1\} \quad (7)$$

Where the numerator gives the dot product and the denominator is the product of the norms of the two vectors $X^i$ and $X^j$, transformed by the inverse of Cholesky decomposed lower triangular matrix of the respective population covariance matrix, $\Lambda_{y=k}^{-1}$.

When the population covariance matrix is known, then doing the vector transformation of the data from sample space to the Euclidean space is quite straightforward. However, one of the objectives of using models containing cosine similarity (like KNN) is to classify test dataset. When population covariance is unknown, problem is faced in test dataset regarding how to perform vector space transformation of test data? Since we do not know the labels of test data beforehand, we do not know whether to apply $\Lambda_{y=1}^{-1}$ or $\Lambda_{y=-1}^{-1}$ to a test data point. Hence, we propose that the test data be transformed into a pseudo-Euclidean space using expected value of $\Lambda^{-1}$.

In the Euclidean space, $X^{Euclidean}$ can be written as

$$X^{Euclidean} = (p)\ X^{Euclidean} + (1-p)\ X^{Euclidean}$$

$$= (p)\ \Lambda_{y=1}^{-1} X^{Random} + (1-p)\ \Lambda_{y=-1}^{-1} X^{Random} \qquad (8)$$

For a test data point, if the probability of it belonging to y=1 is p, then the probability of it belonging to y=-1 is 1-p. Hence the transformation into the Euclidean space will be

$$((p)\ \Lambda_{y=1}^{-1} + (1-p)\ \Lambda_{y=-1}^{-1}) X^{Random} = E(\Lambda^{-1}) X^{Random} \qquad (9)$$

Where $E(\Lambda^{-1})$ is the expected value of $\Lambda^{-1}$. Hence in the absence of information on which sample space a test data point belongs to, we propose to use $E(\Lambda^{-1})$ using the $\Lambda^{-1}$ of the training data to perform transformation to Euclidean Space. Moreover, in absence of population covariance matrix $\Sigma$, we propose to use the sample covariance matrix S and its corresponding Cholesky decomposition:

$$S = W W^T$$

To obtain $E(W^{-1})$ and thus perform the vector transformation from sample space to Euclidean Space.

Value of p is obtained by using the MLE estimates from the training data:

$$p = \frac{n_{y=1}}{n_{y=1} + n_{y=-1}} \qquad (10)$$

where,

$n_{y=1}$= Number of sample observations that have been classified as y=1

$n_{y=-1}$= Number of sample observations that have been classified as y=-1

## 4 CASE STUDY: CLASSIFICATION OF BREAST CANCER DATA

The **Breast Cancer Wisconsin (Diagnostic) dataset**, obtained from the University of Wisconsin Hospitals, Madison from Dr. William H. Wolberg, is a renowned collection of data used extensively in machine learning and medical research. Originating from digitized images of fine needle aspirates (FNA) of breast masses, this dataset facilitates the analysis of cell nuclei characteristics to aid in the diagnosis of breast cancer. The dataset consists of 569 instances of breast cancer diagnoses each having 30 real-valued features identified from the FNA images, with each diagnosis having a class label as Malignant (M) and Benign (B). The 30 features measure various attributes of FNA like radius, texture, perimeter, area, smoothness, compactness concavity, symmetry, fractal dimension, etc.

The objective of this study is to predict the labels of breast cancer observations using the variance adjusted cosine distance and compare its performance with original cosine formula. For this purpose, the dataset was split into training and validation data in the ratio 80:20. Three cases have been analyzed. In the first case the cosine distance of the original input data vectors was calculated and KNN carried out. In the second case the population covariance matrix of all datapoints classed "B" and "M" were calculated separately and the respective $\Lambda_B^{-1}$ and $\Lambda_M^{-1}$ were calculated, then the data points marked "B" or "M" were multiplied with their respective $\Lambda^{-1}$ to perform vector transformation to Euclidean Space, after which cosine distance of the transformed vectors was calculated and KNN was carried out. In the third case the sample covariance matrices of the training data $S_B$ and $S_M$ were calculated and after Cholesky decomposition, the expected Cholesky factor $E(W^{-1})$ was calculated using equation (7). In this case both the training and validation data were transformed to the pseudo-Euclidean space using $E(W^{-1})$ as transformation matrix, cosine distance was calculated in the new transformed space and KNN was applied. Classification table that gives the precision, recall and F1 score for each of the cases was calculated and compared to study respective model performances. Further, leave-one-out cross validation (LOOCV) and k-fold cross validation were carried out on the three cases and the mean accuracy was analyzed and compared.

## 5 Results and Discussion

First before applying KNN, the optimal number of Ks was decided for the base case of KNN application with original cosine similarity formula. Comparing misclassification error vs K gave optimal K =13:

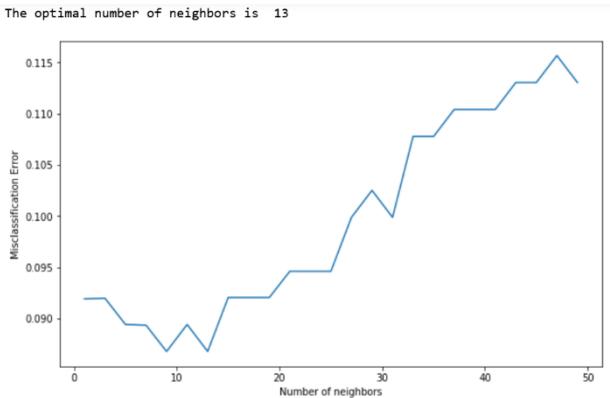

Fig 2: Determination of optimal K

When KNN model is applied using the original cosine distance on the data in original random variable space, the following classification table is obtained:

|  | precision | recall | f1-score | support |
|---|---|---|---|---|
| 0 | 0.921 | 0.986 | 0.952 | 71 |
| 1 | 0.974 | 0.860 | 0.914 | 43 |
| accuracy |  |  | 0.939 | 114 |
| macro avg | 0.947 | 0.923 | 0.933 | 114 |
| weighted avg | 0.941 | 0.939 | 0.938 | 114 |

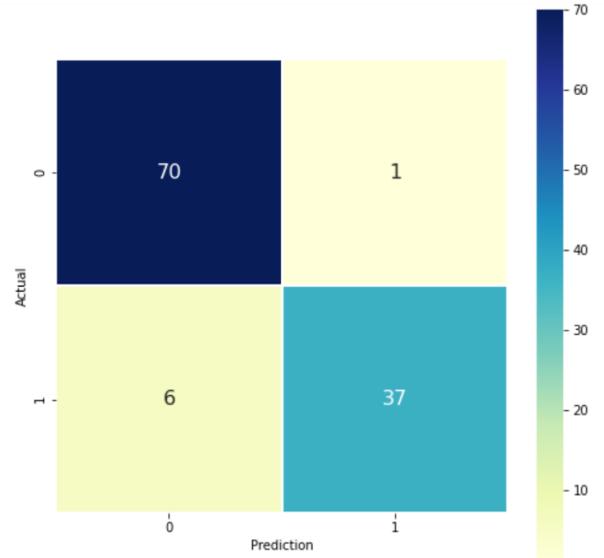

Fig 3: Classification table and Confusion matrix for KNN using original cosine distance

Then the KNN model is applied after data was transformed from the random variable space to the Euclidean space by first taking the population covariance matrix of each class, and then applying the inverse of Cholesky decomposed lower triangular matrix of population covariance matrix of the respective class on the class data. The following classification table is obtained:

|  | precision | recall | f1-score | support |
|---|---|---|---|---|
| 0 | 1.000 | 1.000 | 1.000 | 71 |
| 1 | 1.000 | 1.000 | 1.000 | 43 |
| accuracy |  |  | 1.000 | 114 |
| macro avg | 1.000 | 1.000 | 1.000 | 114 |
| weighted avg | 1.000 | 1.000 | 1.000 | 114 |

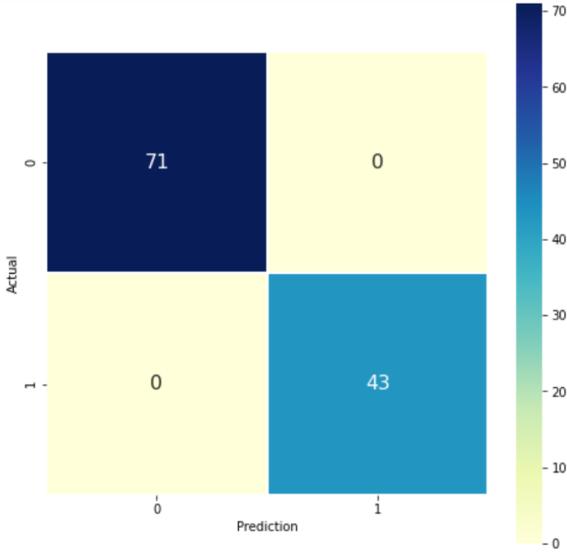

Fig. 4: Classification table and Confusion matrix for KNN using each class population covariance matrix adjusted cosine distance

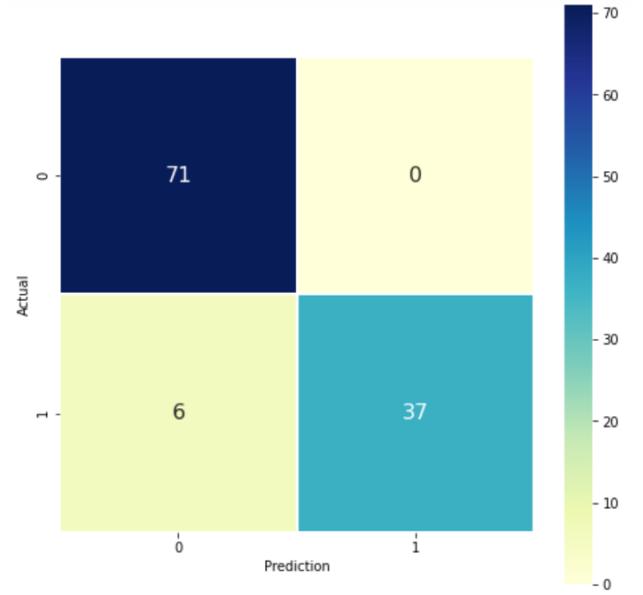

Fig. 5: Classification table and Confusion matrix for KNN using $E(W^{-1})$ adjusted cosine distance.

**Here it can be seen that when the data points are transformed using $\Lambda_{y=1}^{-1}$ or $\Lambda_{y=-1}^{-1}$, it results in 100% test accuracy.** This shows that the space to which the data points are transformed is the Euclidean space and that cosine distance is an accurate measure of similarity in that space. It also explains why cosine similarity acts well when data is normalized, because normalization removes some of the variance effects. However along with adjusting for variance, the data also needs to be decorrelated. Hence $\Lambda^{-1}$ is a better metric.

Due to the limitation of not knowing the population covariance matrix and lack of information regarding which transformation matrix to be used to transform the test data points, in the third case the expected $W^{-1}$ has been used to transform the data. When $E(W^{-1})$ is used for transformation, then the following classification table is obtained:

```
              precision    recall  f1-score   support

           0      0.922     1.000     0.959        71
           1      1.000     0.860     0.925        43

    accuracy                          0.947       114
   macro avg      0.961     0.930     0.942       114
weighted avg      0.951     0.947     0.946       114
```

Here it can be seen that when the expected value $E(W^{-1})$ is calculated in line with (9) it results in higher accuracy compared to applying KNN on un-transformed data in the random variable space.

Finally, two cross validation approaches were tried: leave one out cross validation (LOOCV) and K-fold cross validation. Here we have used 5-fold cross validation. Calculating the mean accuracies of the two validation approaches in the three cases is shown in the table below:

| Table 1: Comparison of Validation approaches | | | |
|---|---|---|---|
| | **Mean Accuracy** | | |
| | KNN-Raw Data | KNN- each class population covariance matrix adjusted cosine distance | KNN-$E(W^{-1})$ adjusted cosine distance |
| Leave-one-out validation | 0.9121 | 1.0 | 0.9244 |
| 5-Fold cross validation | 0.9068 | 1.0 | 0.9191 |

Here we see that in both the cross-validation approaches, KNN on data transformed using each class $\Lambda^{-1}$ has 100% accuracy, and data transformed using $E(W^{-1})$ has higher accuracy than original cosine distance.

## 6 CONCLUSIONS AND FUTURE WORK

The perfect test accuracy in the second case shows that cosine distance is an appropriate indicator of similarity if the data is whitened. This study also shows that doing cosine similarity in the original space in which the data resides (random variable or input space) runs the risk of mis-classification. While this is not a concern if the data is displaying low covariance, in certain situations like text data where significant covariance is observed between certain tokens, the modified cosine distance can be advantageous.

It can be seen that when data is transformed into the Euclidean space using their respective $\Lambda_{y=1}^{-1}$ or $\Lambda_{y=-1}^{-1}$, it results in 100% accuracy in both LOOCV and 5-fold cross validation. Also it can be seen that by transforming the data into a pseudo-Euclidean space with $E(W^{-1})$ results in higher accuracy compared to KNN on untransformed data. Hence in the absence of knowledge of the population covariance matrix, $E(W^{-1})$ is a better metric to do KNN.

Despite the performance improvement as shown in the study, the modified cosine distance suffers from some drawbacks. The primary assumption is that the population covariance matrix for both the classes of data be known. However, in many studies only the sample covariance matrix is known. While an attempt has been made here to use the $E(\Lambda^{-1})$, there is scope of further improvement. Hence future course of work will involve finding a better estimate of the population covariance matrix so that more accurate transformation of the raw data can be carried out.

## 7 DISCLAIMER

The results from this research provided in this manuscript are the sole responsibility of the authors and do not reflect the views or opinions of the institutions where they work.

## 8 DECLARATION OF COMPETING INTEREST

The authors declare that they have no known competing financial interests or personal relationships that could have appeared to influence the work reported in this paper.